\documentclass[12pt,letterpaper]{article}
\usepackage{etoolbox}

\AtBeginEnvironment{tabular}{\scriptsize}

\usepackage[margin=1in]{geometry}
\usepackage{hyperref}       % hyperlinks
\usepackage{url}            % simple URL typesetting
\usepackage{booktabs}       % professional-quality tables
\usepackage{amsfonts}       % blackboard math symbols
\usepackage{nicefrac}       % compact symbols for 1/2, etc.
\usepackage{microtype}      % microtypography
\usepackage{graphicx}
\usepackage{titling}
\PassOptionsToPackage{numbers, compress}{natbib}
\usepackage{amsmath}
\usepackage{mathtools}

\usepackage{color}
\usepackage{multirow}
\usepackage[ruled,vlined]{algorithm2e}
\usepackage{subfig}

\author{\large Michael Minyi Zhang$^{1} $, Bianca Dumitrascu$^{2}$, Sinead A. Williamson$ ^{3} $,\\ 
	\large Barbara E. Engelhardt$ ^{4} $}
\title{Sequential Gaussian Processes for Online Learning of Nonstationary Functions}
\date{\texttt{mzhang18@hku.hk, bmd2151@columbia.edu,  sinead@austin.utexas.edu, barbarae@stanford.}\\
	$^{1}$ Department of Statistics and Actuarial Science, University of Hong Kong.\\
	$^{2}$ Department of Statistics and the Herbert and Florence Institute for Cancer Dynamics, Columbia University.\\
	$^{3}$ Department of Statistics and Data Science, The University of Texas at Austin.\\
	$^{4}$ Department of Biomedical Data Science, Stanford University.\\
	\vspace{2em} 
	\today}	

\DeclareMathOperator*{\argmax}{\arg\!\max}

\newcommand{\calG}{\mathcal{G}}
\newcommand{\calH}{\mathcal{H}}
\newcommand{\I}{\mathbf{I}}
\newcommand{\jast}{\mathbf{j}^{\ast}}
\newcommand{\iksum}{i^{\prime} : (z_{i^{\prime}} = k, i^{\prime} < i)}
\newcommand{\iprime}{i^{\prime}}
\newcommand{\m}{\mathbf{m}}
\newcommand{\N}{\mathcal{N}}
\newcommand{\oneT}{_{1:i}}
\newcommand{\oneTminus}{_{1:i-1}}
\newcommand{\bigO}{\mathcal{O}}
\newcommand{\Real}{\mathbb{R}}
\newcommand{\bu}{\mathbf{u}}
\newcommand{\bs}{\mathbf{s}}
\newcommand{\bS}{\mathbf{S}}
\newcommand{\T}{\mathcal{T}}

\newcommand{\wJ}{w^{(j)}}
\newcommand{\wtJ}{w^{(j)}_{i}}
\newcommand{\wtminusJ}{w^{(j)}_{i-1}}

\newcommand{\Ws}{w_i^{(1)}, \ldots , w_i^{(J)}}
\newcommand{\X}{\mathbf{X}}
\newcommand{\x}{\mathbf{x}}

\newcommand{\y}{\mathbf{y}}

\newcommand{\z}{\mathbf{z}}
\newcommand{\bmu}{\boldsymbol\mu}
\newcommand{\bpi}{\boldsymbol\pi}
\newcommand{\bSigma}{\boldsymbol\Sigma}
\newcommand{\SigmaX}{\Sigma_{\X\X^{\prime}}}
\newcommand{\SigmaXStar}{\Sigma_{\X^{\ast}\X^{\ast\prime}}}
\newcommand{\SigmaXX}{\Sigma_{\X^{\ast}\X}}
\newcommand{\SigmaXXT}{\Sigma_{\X\X^{\ast}}}

\newcommand{\btheta}{\boldsymbol\theta}
\newcommand{\bPsi}{\mathbf{\Psi}}
\newcommand{\uJ}{^{(j)}}

\begin{document}
	\maketitle
	\begin{abstract}
		Many machine learning problems can be framed in the context of estimating functions, and often these are time-dependent functions that are estimated in real-time as observations arrive. Gaussian processes (GPs) are an attractive choice for modeling real-valued nonlinear functions due to their flexibility and uncertainty quantification. However, the typical GP regression model suffers from several drawbacks: 1) Conventional GP inference scales $\bigO(N^{3})$ with respect to the number of observations; 2) Updating a GP model sequentially is not trivial; and 3) Covariance kernels typically enforce stationarity constraints on the function, while GPs with non-stationary covariance kernels are often intractable to use in practice. To overcome these issues, we propose a sequential Monte Carlo algorithm to fit infinite mixtures of GPs that capture non-stationary behavior while allowing for online, distributed inference. Our approach empirically improves performance over state-of-the-art methods for online GP estimation in the presence of non-stationarity in time-series data. To demonstrate the utility of our proposed online Gaussian process mixture-of-experts approach in applied settings, we show that we can sucessfully implement an optimization algorithm using online Gaussian process bandits.
% this last sentence kind of comes out of nowhere. what is "our MoE approach" and what is "bandit optimization"?
	\end{abstract}
	\section{Introduction}\label{sec:intro}
Data are often observed as streaming observations that arrive sequentially across time. Examples of streaming data include hospital patients' vital signs, telemetry data, online purchases, and stock prices. To model streaming data, it is more efficient to update model parameters as new observations arrive than to refit the model from scratch with the new observations appended onto existing data. A typical prior distribution on the space of functions for time series analysis is the Gaussian process \cite{Rasmussen:Williams:2006}. Gaussian processes (GPs) are a convenient distribution on real-valued functions because, when evaluated at a fixed set of inputs, they have a multivariate normal distribution and hence allow closed-form posterior inference and prediction when used for regression. 

Parameter estimation for GPs remains challenging because inference involves calculating the Gaussian likelihood. This means that the computational complexity of inference is dominated by the $\bigO(N^{3})$ operation of inverting an $ N\times N $ matrix, where $ N $ is the number of observations. This complexity makes standard GP inference techniques unsuitable for scenarios with large numbers of observations or where fast inference is essential. Further, Bayesian approaches, such as Markov chain Monte Carlo (MCMC) or variational inference (VI), do not trivially allow online updating of model parameter estimates, although sequential Monte Carlo methods (SMC) may be adapted for this purpose. 

From a statistical perspective, a typical GP regression model infers only stationary functions, meaning that properties of the function are constant across all input values. While there are covariance kernels that explicitly capture non-stationary effects in GP regression, they pose greater computational challenges than a stationary kernel as they often require calculating intractable integrals. Mixture-of-experts GP models have been used to model non-stationary functions by fitting independent GPs to different segments of the input space.  In particular, the importance sampled mixture-of-experts (IS-MOE) approach fits mixtures of GP experts in a distributed manner using importance sampling \cite{Zhang:Williamson:2017}; however, IS-MOE is not an online algorithm. Conversely, sparse online GPs are a state-of-the-art method for online GP estimation, but the estimated functions are constrained to be stationary. 

Given the prevalence and complexity of streaming data, this gap necessitates a new approach for online learning of non-stationary functions. Moreover, in most mixture-of-experts approaches, we assume the number of mixtures are fixed and known \textit{a priori}, which is generally inappropriate. Instead, we can assume that there is an infinite mixture of GPs, by assuming the mixing distribution is Dirichlet-process distributed. We introduce a sequential algorithm to fit infinite mixtures of GPs. SMC samplers can be adapted to allow real-time updates to the model parameters, and are trivially parallelizable. We show a connection with online GPs to multi-armed bandits for optimization and demonstrate that our method can obtain superior performance compared to other GP-bandit optimization techniques.  

This paper proceeds as follows: In Section~\ref{sec:background}, we discuss the background for Gaussian processes, methods for fast inference, and importance sampling. We introduce our online GP inference algorithm in Section~\ref{sec:method}. We show empirical benefits of our framework on prediction tasks in both simulated data and in hospital patient data where online, scalable inference is essential in Section~\ref{ref:experiments}. In Section~\ref{sec:conclusion}, we conclude with a discussion of future work.

	\section{Problem Statement and Background}\label{sec:background} 
\subsection{Gaussian Processes}\label{subsec:background:gp}
Consider the problem of estimating an unknown function $f: \Real^D \rightarrow \Real$ from streaming data. In detail, we have $ D $-dimensional input data $\x_i \in \Real^D$, and output data, $ y_i \in \Real $,  which arrive at times $i = 1,2, \ldots , N$. We assume a functional relationship, $y_i = f(\x_i) +\epsilon_i $, where $\epsilon_i \sim \mathcal{N}(0,\sigma^2)$ where the function $ f $ is assumed to be distributed from a Gaussian process. Gaussian processes (GPs) are a popular approach to modeling distributions over arbitrary real-valued functions $f(\cdot)$, with applications in regression, classification, and optimization, among others. Characterized by a mean function $\mu(\cdot)$, and a covariance function $ \Sigma(\cdot,\cdot)$, we can sample instantiations from a GP at a fixed set of locations $\X = [\x_1, \dots , \x_N]^{T}$, according to an $N$-dimensional multivariate normal:
\begin{equation*}
f(\X)|\X \sim \N_N(\mu(\X), \SigmaX),
\end{equation*}
where $\mu(\X)$ and $\SigmaX$ are the mean and covariance functions evaluated at the input data $\X$. Typically, the mean function is set to be zero in Gaussian process regression. In the presence of noisy observations, $\y = [y_1, \ldots , y_N]^{T}$, whose mean value is a GP-distributed function, we observe the data generating process: 
\begin{equation}
\y|\X,f \sim \N_N \left(f(\X),\sigma^2 \I\right), \hspace{1em} f|\X \sim \N\left(0,\SigmaX \right).
\end{equation}

Due to conjugacy between the likelihood and prior, we can marginalize out $ f $ analytically such that:
\begin{equation}
\y | \X \sim \N_N\left( 0 , \SigmaX  + \sigma^{2} \I \right).
\end{equation}
We typically fit a GP regression model by optimizing the marginal likelihood with respect to the model hyperparameters. Given some previously unseen inputs $ \X^{\ast}  $, the posterior predictive distribution of the outputs $ \y^{\ast} $ is
\begin{equation}
\begin{aligned}
&\m_{\y^{\ast}} =  \SigmaXX \SigmaX^{-1} \y\\
&\bS_{\y^{\ast}} = \SigmaXStar -  \SigmaXX \SigmaX^{-1}\SigmaXXT + \sigma^{2} \I \\
&\y^{\ast} | \y, \X, \X^{\ast} \sim \N \left(\m_{\y^{\ast}}, \bS_{\y^{\ast}} \right).
\end{aligned}
\end{equation}

\subsection{Fast Gaussian Process Inference}\label{subsec:background:gp_inference}
Fitting GP-based models is dominated by $ \bigO(N^{3}) $ covariance matrix inversion operations, meaning that these models are challenging to fit to large sample size data. To allow GP models to be computationally tractable for large data, numerous approaches have been developed for scalable GP inference. These approaches largely fall into two groups: sparse methods and local methods. Sparse methods approximate the GP posterior distribution with an $ M \ll N $ number of inducing points that act as pseudo-inputs to the GP function, thereby reducing the computational complexity to $ \bigO(NM^{2}) $ \cite{Snelson:Ghahramani:2005,Titsias:2009}. We can further speed up these sparse methods using stochastic variational inference (SVI) that reduces the complexity by fitting the model with subsets of the $ M $ inducing points to form stochastic gradients at each iteration \cite{Hensman:2013}. 

Local methods, on the other hand, exploit structure among samples to represent the covariance matrix using a low-rank approximation \cite{Deisenroth:Ng:2015,Cohen:Mbuvha:Marwala:Deisenroth:2020}. To do this, they partition the data into $ K $ sets, and approximate the covariance matrix by setting the inter-partition covariances to zero. They invert this approximate covariance matrix by inverting $ K $ dense matrices of size $ N/K $, resulting in a complexity of $ \bigO( N^{3}/K^{2}) $. An additional benefit of local approaches is that we can estimate GP hyperparameters within each block; when samples are partitioned based on input location, this implicitly captures non-stationary functions. A product of experts model implies the strong assumption that, across experts, the observations are independent. This independence assumption ignores the correlation between experts and can lead to poor posterior predictive uncertainty quantification. 

Instead, a more flexible approach to local methods in the GP domain is to assume that there is a mixture of GP-distributed experts.  Mixture-of-experts GP models can model non-stationary functions \cite{Gramacy:Lee:2005,Rasmussen:Ghahramani:2002,Meeds:Osindero:2006,Tresp:Mixtures:2000}, but integration over partitions makes inference computationally intractable for most realistic settings. In contrast, variational methods for mixtures of GPs are effective to speed up computation \cite{Yuan:Neubauer:2009, Sun:2010, Nguyen:Bonilla:2014}. %present an approximate variational method to speed up computation for mixtures of GPs. 
Alternate methods using sum-product networks and importance sampling have also proven to be fast and effective for Bayesian inference of GP mixtures \cite{Trapp:Peharz:Pernkopf:Rasmussen:2020,Zhang:Williamson:2017}. %introduced faster exact methods for through . 

\subsection{Online Gaussian Process Inference}\label{subsec:background:smc}
The online product-of-experts (POE) variant of GP regression assumes the data are strictly partitioned--leading to a block diagonal structure in the covariance matrix \cite{Nguyen:Peters:Seeger:2009}. Contrast this to the mixture-of-experts approach, %and our proposed method, 
where the partition is integrated out, leading to a more expressive covariance structure, but is not inherently amenable to online updating.

Although local methods are simple to update in real-time, as we can send new data to clusters and only update clusters receiving new data, sparse and mixture-of-experts GP inference methods discussed above do not trivially allow model updates as new data arrive. Previous work in online GP inference required non-trivial adaptations to allow real-time GP inference. The sparse online Gaussian process is an inducing-point online method by approximating the posterior using variational inference. However, this approach assumes the kernel hyperparameters are fixed and known \textit{a priori} \cite{Csato:Opper:2002}. The online sparse variational Gaussian process (OSVGP) is another inducing point variational method for online GP regression that updates the hyperparameters as new data arrive \cite{Bui:Nguyen:Turner:2017}. 

Many scalable inference algorithms take advantage of training the model only using a small subset of the data in a ``minibatching'' approach. Examples of minibatches being used for scalable inference include stochastic gradient descent \cite{Robbins:Monro:1951}, stochastic variational inference (SVI) \cite{Hoffman:2013} and stochastic gradient MCMC \cite{WelTeh2011}. However, in the sparse online settings we cannot use minibatching in a stochastic approximation setting. %As discussed by, 
This is because SVI requires minibatches uniformly sampled \textit{i.i.d.} from the entire dataset. But in the streaming case, samples from the new data may not be from the same distribution as the previous observations \cite{Bui:Nguyen:Turner:2017}. Moreover, minibatching in an SVI-type setting requires constant updating of the entire model, unlike in local methods that can take minibatches of the incoming data and only update a small subset of experts.

%In special cases, we may take advantage of a Kronecker structure in the covariance kernel whereby 
One notable variant of exact inducing point online method for streaming GP regression, called the Woodbury inversion with structured kernel interpretation (WISKI)  \cite{Stanton:Maddox:Delbridge:Wilson:2021}, replaces the covariance kernel with a structured, sparse matrix approximation for GP regression \cite{Wilson:Nickisch:2015}. Using this ``structured kernel interpolation'' method in conjunction with the Woodbury rank-one updates to the kernel inverse leads to an online GP method with constant computational complexity with respect to the number of observations. Empirically, %we observe that 
the online variational method of OSVGP
%Additionally, we can see in  %\cite{Stanton:Maddox:Delbridge:Wilson:2021} note that the online variational GP method by \cite{Bui:Nguyen:Turner:2017} 
tends to suffer from numerical stability issues and is vulnerable to underfitting the model in the streaming setting \cite{Stanton:Maddox:Delbridge:Wilson:2021}. Alternatively, we may interpret sparse online GP under  a different formulation. %of sparse online GP regression. 
Instead of the typical approaches in the aforementioned papers, the authors reframe the sparse model as a state space model and use the Kalman filtering update equations to train the model sequentially \cite{Schurch:2020}.

\subsection{Modeling Non-Stationary Functions with Gaussian Processes}
Modeling non-stationary functions using GPs is often computationally challenging because non-stationary kernels generally include more parameters than stationary kernels and are more computationally intensive to fit. %, and many non-stationary kernels require calculating intractable integrals. %  \cite{Higdon:Swall:Kern:1999}. 
Many non-stationary kernels model the hyperparameters as a GP-distributed function of the input space \cite{Kersting:Plagemann:Pfaff:Burgard:2007,Tolvanen:Jylanki:Vehtari:2014,Heinonen:Mannerstrom:2016,Paciorek:2006}. In this setting, if we have a GP-distributed kernel hyperparameter for each observation then the computational complexity of posterior sampling the hyperparameters scales $ \bigO(N^{4}) $ as we have to update $ N $. 
%and therefore incur $\bigO(D^3)$ costs to learn the posterior which generates the kernel in addition to the $\bigO(N^3)$ cost of learning hyperparameters . 

The local Gaussian process product-of-experts approach to fitting online GPs that can account for non-stationary behavior, but inference depends on a single partitioning of the input data using $K$-means \cite{Nguyen:Peters:Seeger:2009}. Hard partitioning may create undesirable edge effects due to its implicit assumptions that partitions have zero correlation. The Bayesian treed GP addresses the problem of hard partitioning by integrating over the space of decision tree partitions via reversible jump MCMC \cite{Gramacy:Lee:2005}. While this method flexibly captures non-stationary functions, reversible jump is slow to marginalize over the space of treed GP partitions and is not parallelizable due the Markov dependencies in MCMC.

The Gaussian process change point model captures non-stationary behavior in GPs by jointly modeling a GP regression model and a change point generating process \cite{Saatci:Turner:Rasmussen:2010}. The prior distribution on the change point locations is assumed to follow a survival function that is known \textit{a priori}, which represents the run time of a functional regime. This model estimates the change point runtime in an online setting, based on a Bayesian online change point detection model \cite{Adams:MacKay:2007}. 

\subsection{Dirichlet Process Mixture Modeling}
For a finite mixture model, we assume that there are $ K $ mixtures where the observations $ \y $ are parameterized by $ \theta_k \sim \calH$ for some distribution $ \calH $ defined on a parameter space $\Theta$ with mixture weights $ \bpi = [\pi_1, \ldots , \pi_K]  $, where $ 0 < \pi_k < 1$ and $ \sum_{k=1}^{K}\pi_k = 1 $:
\begin{equation}
P(\y|\btheta, \bpi) = \prod_{i=1}^{N}\sum_{k=1}^{K} \pi_k P(y_i | \theta_{k}). 
\end{equation}
For the purpose of computational tractability, we can augment the mixture model with a latent indicator $ z_i \in \left\{ 1, \ldots , K \right\} $, which tells us the latent mixture membership for observation $ y_i $, so that our mixture model is now parameterized as:
\begin{equation}
\begin{aligned}
&P(\y|\btheta, \z, \bpi,\alpha) = \prod_{i=1}^{N}  P(y_i | \theta_{z_i}), \hspace{.5em} P(z_i | \bpi) \sim \mbox{Categorical}(\bpi) \\
&P(\bpi | \alpha) \sim \mbox{Dirichlet}\left( \frac{\alpha}{K}, \ldots , \frac{\alpha}{K} \right), P(\theta_k) \sim \calH.
\end{aligned}
\label{eqn:dpmm}
\end{equation}

In our proposed approach, we assume that now there is an \textit{a priori} infinite number of mixtures and that the mixture distribution is generated from a Dirichlet process. The Dirichlet process (DP) is a distribution over probability measures $\calG \sim \mbox{DP}(\alpha, \calH)$, defined on some parameter space, $\Theta$, with a concentration parameter $ \alpha \in \Real^{+} $. We define a Dirichlet process by its finite dimensional marginals. For a finite partition of $ \Theta $, $\left( A_1, \ldots , A_K \right)$, if the probability masses $\calG(A_1), \dots, \calG(A_K)$ are $\mbox{Dirichlet}\left(\alpha \calH(A_1), \dots, \alpha \calH(A_K)\right)$-distributed then $\calG\sim \mbox{DP}(\alpha, \calH)$  \cite{Ferguson:1973}. We obtain the Dirichlet process mixture model (DPMM) by taking the limit in Equation~\ref{eqn:dpmm} as $ K\rightarrow\infty $. 

Furthermore, we can integrate out the mixing proportion $ \bpi $ and sample the latent indicators $ \z $ from the predictive distribution of the Dirichlet process--the Chinese restaurant process \cite{Aldous:1985,Pitman:2002}. In the Chinese restaurant process (CRP), the $ i $-th latent indicator probability is now:
\begin{equation}
P(z_i = k | z_1, \ldots , z_{i-1}) = \left\{ \begin{array}{ll} 
\frac{N_k^{\prime}}{i-1+\alpha} &k \in \left(z_1, \ldots , z_{i-1}\right), \\ 
\frac{\alpha}{i-1+\alpha} &\mbox{otherwise,}
\end{array} \right.
\end{equation}
where $ N_k^{\prime} = \sum_{i^{\prime}=1}^{i-1}I(z_{i^{\prime}}=k) $ is the number of previous observations from $ z_1 $ to $ z_{i-1} $ assigned to mixture $ k $.

We place this non-parametric assumption on the mixture distribution because we expect there will be an infinite number of mixtures as we observe an infinite amount of data, but the DPMM will adaptively instantiate a finite number of mixtures for a finite number of observations. 

\subsection{Importance Sampling and Sequential Monte Carlo}
A central issue in Bayesian inference is the computation of the often intractable integral $ \bar{f} = \int f(\theta)P(\theta)  d\theta$. For this task, importance sampling (IS) is a popular tool. To perform importance sampling, $ J $ samples of $\theta\uJ$ (also known as ``particles'') are drawn from a proposal distribution $ Q(\theta) $ that is simple to sample from and approximates $P(\theta)$ well. The integral is estimated using the weighted empirical average of these samples $ f(\theta\uJ) $ with weights $\wJ =  P\left( \theta\uJ\right)/Q\left(\theta\uJ\right) $ to approximate the integral:
\begin{equation}
\int f(\theta)P(\theta) d\theta \approx  \sum_{j=1}^{J} \wJ f \left(\theta\uJ\right).
\end{equation}
For problems where $ \theta $  arrives in a sequence, $ \btheta_{1:N} = (\theta_{1}, \ldots , \theta_{N})$, we can rewrite $\wJ$ as a sequential update of the importance weight:
\begin{equation}
\wtJ = \wtminusJ \frac{   P( \btheta\oneT )  }{ P(\btheta\oneTminus)  Q(\theta_{i} | \btheta\oneTminus)  }.\label{eqn:sis_weight}
\end{equation}
In a Bayesian context, one can view this sequential importance sampler as sequentially updating a prior distribution at $i=0$ to the posterior of all observed data at time $N$. Moving from $ i=1 $ to $ N $ in this sampler may lead to a situation where the weight of one particle dominates the others, thereby increasing the variance of our Monte Carlo estimate and propagating suboptimal particles to future time steps. To avoid this, we can resample the particles according to their weights $ \wJ $ for $ j=1, \ldots , J $ particles--leading to the sequential Monte Carlo (SMC) sampler \cite{DelMoral:Doucet:Jasra:2006}. Monte Carlo techniques like IS and SMC are appealing from a computational perspective as we can trivially parallelize the computation for each weight $ \wtJ $ to as many processors available as each weight is independent of the other.
%In the Bayesian context, our target distribution is the posterior:
%\begin{equation}
%P(\btheta_{1:i} | \X_{1:i}) = \frac{P(\X_{1:i} | \btheta_{1:i})P(\btheta_{1:i})}{ P(\X_{1:i})} .
%\end{equation}
%If the proposal distribution for $ \theta_i $ that we sample from is an MCMC sample from $ P(\btheta_{1:i} | \X_{1:i}) $ then our proposal weight simplifies to the following expression:
%\begin{equation}
%\wtJ = \wtminusJ \cdot \frac{P(\X_{1:i} | \btheta_{1:i})P(\btheta_{1:i})}{P(\X_{1:i-1} | \btheta_{1:i-1})P(\btheta_{1:i-1})}
%\end{equation}

We can sequentially update a Dirichlet process mixture model for online inference using an SMC sampler under general proposal distributions and modeling assumptions \cite{Ulker:Gunsel:Cemgil:2010}. %introduce a SMC method for DP mixtures with a general proposal distribution. 
In the particle filtering algorithm for DP mixtures, we have an analytically convenient expression for the weight updates in the special case where the observation model is a Dirichlet process of normal inverse-Wisharts \cite{Carvalho:Lopes:Polson:Taddy:2010}. Moreover, SMC methods have also been used in online GP learning %For example, \cite{Gramacy:Polson:2011} propose an SMC sampler 
for sequentially updating GP hyperparameters, but rely on strong modeling and conjugate prior choices as this SMC sampler takes advantage of closed-form marginalization of nuisance parameters for efficient inference \cite{Gramacy:Polson:2011}. Later work has generalized the SMC sampler for updating the GP hyperparameters sequentially without restrictive conjugacy requirements \cite{Svensson:Dahlin:Schon:2015}. However, these methods are not truly online methods and are unsuitable for large-scale streaming scenarios because the complexity of updating the particle weight still grows cubically with respect to the number of observations.

Finally, IS-MOE is a fast method that unifies non-stationary function learning and parallel GP inference \cite{Zhang:Williamson:2017}. To achieve scalability, IS-MOE integrates over the space of partitions using importance sampling, which allows distributed computation. This method uses ``minibatched'' stochastic approximations, where the model is fit with only a subset of the training set and the likelihood is upweighted to approximate the full data likelihood. However, IS-MOE cannot update GP models as new data arrive. For this purpose, we develop a new approach for online, parallelizable inference of a mixture-of-experts GP model that we call ``GP-MOE''. % to be computationally tractable for streaming data.

\subsection{Gaussian Processes and Optimization: A Bandit Perspective}
Bayesian optimization techniques seek to find parameters %$\btheta$ 
that best model a conditional probability. %$P(\y|\X,\btheta)$. 
Many approaches optimize parameter configurations adaptively \cite{Hutter:2011, Bergstra:Bardenet:Bengio:Kegl:2011,Snoek:Larochelle:Adams:2012}, with bandit formulations being particularly successful in GP contexts \cite{Wang:Zoghi:Hutter:co:2013, Srinivas:Krause:Kakade:Seeger:2009,li2016hyperband}. The bandit framework considers the problem of optimizing a function $f$, sampled from a GP, 
%with hyperparameters $\btheta$, 
by sequentially selecting from a set of arms corresponding to inputs $\x_{i}$, where noisy values $y_{i}$ are observed.% and rewards are

The goal of a bandit algorithm is to sequentially select arms that maximize long-term rewards; to do this, one needs to approximate the expected rewards for each arm (exploration) and then select the arms that maximize rewards (exploitation). There are two common algorithmic strategies for choosing arms: optimization based algorithms and sampling based approaches. Here, we look at the context of using GP-distributed functions for bandit optimization. At time $i$, we select a new point $ \x_i $ that maximizes some acquisition function. We can select the next point to evaluate in the function by optimizing with respect to the predictive reward:
\begin{equation}
\x_i := \argmax_{\x^{\ast} \in \Real^{D}}  \m(\x^{\ast}) ,
\end{equation}
which can quickly converge to a poor local maxima. To better explore the function, we can select $ \x_i $ with respect to the GP upper confidence bound (GP-UCB):
\begin{equation}
\x_i := \argmax_{\x^{\ast} \in \Real^{D}}  \m(\x^{\ast}) + \sqrt{\beta_i \cdot \bs(\x^{\ast}) },
\label{eqn:ucb}
\end{equation}
where $ \m(\cdot), \bs(\cdot) $ are the predictive posterior mean and variance for a test point $ \x^{\ast} $, and $ \beta_{i} $ is a tuning parameter that acts as the trade-off between exploration and exploitation. The GP-UCB acquisition function prefers to select new points where there is both high uncertainty and a high predicted reward \cite{Srinivas:Krause:Kakade:Seeger:2009}.

In Thompson sampling (TS), we take samples from the posterior predictive distribution evaluated at $ \X^{\ast} $ and select the test input that yields the highest sample of $ y^{\ast} $:
\begin{equation}
\begin{aligned}
(y^{\ast}_1, \ldots , y^{\ast}_{N_{samp}}) \sim P(\y^{\ast} | \X, \X^{\ast}, \y ),\\
\x_i :=  \x_{\argmax_{i^{\ast}} (y^{\ast}_1, \ldots , y^{\ast}_{N_{samp}})}.
\end{aligned}
\label{eqn:ts}
\end{equation}
The TS approach to the bandit problem is closely related to optimizing the acquisition function with respect to the UCB bound and can be seen as a sampling-based variant balancing the exploration and exploitation trade-off \cite{russo2014}.

%In addition to analyzing how well a bandit can obtain the optimal value of function, a common metric for evaluating bandit performance is the extent to which the bandit can minimize ``regret''--the residual between the predicted value of selected point at time $ i $ and its true value:
%\begin{equation}
%\epsilon_i = f(\x_i) - \m(\x_i).
%\end{equation}

Online GPs are a naturally appealing method for Bayesian optimization. The function being optimized, $ f $, is usually difficult to evaluate. Hence, we are unlikely to observe more than handful of evaluations of $ f $ upon initially fitting the GP model, and we are unlikely to have an accurate estimate of the hyperparameters at the initial fit of the model. Since we do not know the GP hyperparameters of Bayesian optimization \textit{a priori} either, we should update the GP as new function queries arrive.

	\section{Sequential Gaussian Processes for Online Learning}\label{sec:method}

We assume that our data is generated from a Gaussian process mixture, similar to previous mixture-of-expert models for Gaussian processes.  This hierarchical model allows for greater flexibility in modeling functions, at the cost of more difficult inference for which we propose a distributable solution in the next section. Our approach adopts the following generative model: 
\begin{align}
\begin{split}
\x_i \sim \T (\bmu_{z_i}, \bPsi_{z_i}, \nu_{z_i}), \hspace{1em} \alpha\sim\mbox{Gamma}(a_0, b_{0}),\\  
z_i | \alpha \sim \mbox{CRP}(\alpha), \hspace{1em}  (\theta_{k}, \sigma^2_{k}) \sim \log\N(\m_0, s^{2}_{0}\I),\\
\y_{k} | \X_k, \theta_{k}, \sigma^2_{k} \sim \mathcal{N}(0, \bSigma_{\theta_{k}} + \sigma^2_{k} \I),
\label{eqn:ind_gps}
\end{split}
\end{align}
where $ \left(\X_k, \y_{k}\right) = \left( \x_i, y_i : z_{i} = k \right) $ represent the data associated with the mixture $k$. We assume that the inputs are distributed according to a Dirichlet process mixture of normal-inverse Wishart distributions, and we marginalize out the parameter locations from the inputs. 
%If 
%\begin{equation}
%\begin{split}
%\x_i \sim \N(\m_{z_i}, \bS_{z_i}  ), \m_{k} \sim \N(\bmu_0, \lambda_{0}^{-1} \bS_k  ), \\
%\bS_k \sim \Winv(\bPsi_{0} ,\nu_0 ),
%\end{split}
%\end{equation}
%then marginalizing out $ (\m_{k}, \bS_{k}) $ results in a multivariate $ t $ distribution:
%\begin{equation}
%\x_i \sim \T(\bmu_{z_i}, \bPsi_{z_i}, \nu_{z_i}) 
%\end{equation}
The outputs are then assumed to be generated by independent GPs, given the mixture indicator. The GP parameters, $ \btheta_{k} = [\theta_{k}, \sigma^2_{k}] $, are assumed to be log-normally distributed, and we update the state of $ \btheta_{k} $ using the elliptical slice sampler \cite{Murray:Adams:MacKay:2010}. 

The elliptical slice sampler (ESS) is an MCMC technique for sampling from a posterior distribution where the prior is assumed to be a multivariate Gaussian distribution but the likelihood is not conjugate. For the ESS algorithm, the sampler is always guaranteed to accept a transition to a new state unlike typical Metropolis-Hastings samplers and therefore is very efficient for MCMC sampling. In fact, slice sampling techniques are preferable in posterior sampling of the covariance function hyperparameters in GP models \cite{Murray:Adams:2010}. 

We assign the $ i $-th input sequentially to clusters according to the Chinese restaurant prior and the mixture locations marginalized out \cite{Neal:2000}:
\begin{equation}
P(z_{i} = k| \alpha, \X_k) \propto \left\{\begin{array}{ll}
N_k^{\prime} \cdot \T(\bmu_{k}^{\prime}, \bPsi_{k}^{\prime}, \nu_{k}^{\prime}) & k \in K^{+}. \\
%\int  P(X_{-i,j,k} | \mu_k, \Gamma_k) P(\mu_k,\Gamma_k) d\mu_k d\Gamma_k 
\alpha \cdot \T (\bmu_{0}, \bPsi_{0}, \nu_{0} )& \mbox{o.w.}
%\int P(x_i | \mu_k, \Gamma_k) P(\mu_k,\Gamma_k) d\mu_k d\Gamma_k 
\end{array}\right.\label{eqn:marg_t}
\end{equation}
where $\T(\bmu_{k}^{\prime}, \bPsi_{k}^{\prime}, \nu_{k}^{\prime})$ is a multivariate-$t$ distribution with the mean, covariance, and degrees-of-freedom parameters for observation $i$'s sequential assignment to mixture $k$, where $ K^{+} $ refers to the previously occupied clusters, $ \{ k: N_k^{\prime} > 0 \} $.
\begin{equation}
\begin{aligned}
\bmu_{k}^{\prime} = \frac{\lambda_0\bmu_{0} + N_k^{\prime} \bar{\x}_{k}}{\lambda_{k}^{\prime}}, \bar{\x}_{k}^{\prime} = \frac{\sum_{\iksum} \x_{i^{\prime}} }{N_k^{\prime}},\\
N_k^{\prime} = \sum_{i^{\prime}=1}^{i-1}I(z_{i^{\prime}}=k), \lambda_{k}^{\prime} = \lambda_{0} + N_k^{\prime},\\
%\x_{k}^{\prime} = [\x_{i^{\prime}} : \iksum ]^{T} ,\\ 
\nu_{k}^{\prime} = \nu_{0} + N_k^{\prime} - D + 1,\\
\bPsi_{k}^{\prime} = \frac{\lambda_{k}^{\prime}+1}{\lambda_{k}^{\prime}\nu_k^{\prime}} \left(\bPsi_0  + \bS_{k}^{\prime} + \bS_{\bar{\x}_{k}}^{\prime} \right)\\
\bS_{k}^{\prime} = \sum_{\iksum} \left(\x_{i^{\prime}} - \bar{\x}^{\prime}_k \right)\left(\x_{i^{\prime}} - \bar{\x}^{\prime}_k \right)^{T}\\
\bS_{\bar{\x}_{k}}^{\prime} = \frac{\lambda_{0} N_k^{\prime}}{\lambda_{k}^{\prime}} \left(\bar{\x}^{\prime}_k - \bmu_{0}\right)\left(\bar{\x}^{\prime}_k - \bmu_{0}\right)^{T}.
\end{aligned}
\end{equation}
We use the $ (\cdot)^{\prime} $ notation to indicate that the summary statistics are conditioned only on observations $ i^{\prime} =1 , \ldots, i-1 $. %as opposed to all observations, $ i=1, \ldots, N $. 
We also place a Gamma prior on the DP concentration parameter, $ \alpha $, which allows us to easily sample its full conditional up to observation $ i $ with a variable augmentation scheme \cite{Escobar:West:1995}:
\begin{equation}
\begin{aligned}
\rho | \alpha &\sim&&\mbox{Beta}(\alpha +1, i), K = | \{ k : N_k > 0 \} |\\
\frac{\pi_\alpha}{1-\pi_\alpha} &=&&\frac{a_0 + K -1}{N (b_0 - \log\rho) }\\
\alpha | \z_{1:i}, \pi_\alpha, \rho &=&&(1-\pi_\alpha)\cdot\mbox{Gamma}(\alpha_0 + K -1, b_0 -\log\rho)\\
&&&+ \pi_\alpha\cdot\mbox{Gamma}(\alpha_0 + K, b_0 -\log\rho). 
\end{aligned}
\label{eqn:alpha_update}
\end{equation}

\subsection{SMC for Online GP-MOE}
In an SMC setting with $j=1,\ldots , J$ particles, we first propagate the particles $ (\z^{(j)}, \btheta^{(j)}, \alpha^{(j)}) $ from time $ i-1 $ to $ i $ and fit a GP product-of-experts model. Then we calculate the particle weights. At the initial time, $i=1$, the particle weight $ j $ is:
\begin{align}
\begin{split}
w_1^{(j)} &\propto P(y_1 | z_1^{(j)},  \x_1, \btheta^{(j)} )P(\x_1 | z_1^{(j)}, \alpha^{(j)}).
% \propto \frac{P(Z^{(1)}_j|X^{(1)},Y^{(1)})}{P(Z_j^{(1)}|X^{(1)})} \propto P(Y^{(1)}|X^{(1)},Z_j^{(1)})
% &= \prod_{k=1}^{K_j} \int P( Y^{(1)}_{j,k}  | X^{(1)}_{j,k}, \theta_{j,k})P(\theta_{j,k})d\theta_{j,k}.
\end{split}
\label{eqn:importance_weight}
\end{align}
For $ i > 1 $, the particle weight in Equation~\ref{eqn:sis_weight} can written as the product of:
\begin{enumerate}
	\item The previous weight, $ w_{i-1}^{(j)} $,
	\item The ratio of the model's likelihood up to observation $ i $ over the likelihood up to observation $ i-1 $ \cite{Svensson:Dahlin:Schon:2015},
	\item The particle weight of $ z_i^{(j)} $ for the Dirichlet process mixture model \cite{Carvalho:Lopes:Polson:Taddy:2010}.
\end{enumerate}
The GP term of the particle weight from \cite{Svensson:Dahlin:Schon:2015} in this setting simplifies to the ratio of the new likelihood (including observation $ i $) and the old likelihood (excluding observation $ i $) of the mixture $ z_i $. We can store the old likelihood in memory from the last time we updated the particle weights, so the only computationally intensive step is computing the new likelihood for mixture $ z_i $. This particle update is: 
\begin{equation}
%\wtJ = \wtminusJ \cdot \frac{ \prod_{k \in K^{+}} P(\y_k | \X_k, \z_{1:i}^{(j)}, \btheta_{k}^{(j)}) }{\prod_{k\in K^{\prime+}} P(\y_k^{\prime} | \X_k^{\prime}, \z_{1:i-1}^{(j)}, \btheta^{(j)}_{k}) } \cdot P( \x_{i} |z_{i}^{(j)}, \alpha^{(j)} )			
\wtJ := \wtminusJ \cdot \frac{  P\left( \y_{z_{i}^{(j)}} \left| \X_{z_{i}^{(j)}},  \btheta_{z_{i}^{(j)}}^{(j)} \right. \right) }{ P\left(\y_{z_{i}^{(j)}}^{\prime} \left| \X_{z_{i}^{(j)}}^{\prime},  \btheta^{\prime (j)}_{z_{i}^{(j)}}\right.\right) } \cdot P( \x_{i} |z_{i}^{(j)}, \alpha^{(j)} )
\label{eqn:weight_update} 
\end{equation}
where
\begin{equation}
\begin{aligned}
(\X_{k}^{\prime}, \y_{k}^{\prime}) = &(\x_{\iprime}, y_{\iprime} , \iksum ), \\  &(\X_{k},\y_{k}) = (\x_{i},y_{i},  i: z_{i}=k)
\end{aligned}
\end{equation}
After calculating the particle weights, we calculate the effective sample size, $ N_{\mbox{eff}} = 1/ \sum_{j=1}^{J}\left(\wtJ\right)^{2} $. If the effective number of samples drops below a certain threshold (typically $ J / 2 $), then we resample the particles with probability $ \Ws $ to avoid the particle degeneracy problem. The details for updating the particles are in Algorithm~\ref{alg:bandit_online_update}.

To calculate the predictive posterior distribution of the GP-MOE for test data $ \x_{\ast} $, we calculate the predictive mean and variance on each individual particle, averaged over the mixture assignment for the test data. Then, we average the predictive distribution on each particle, weighted by $ \Ws $. The details for prediction in GP-MOE are available in Algorithm~\ref{alg:bandit_online_predict}.

Assuming each batch is on average size $N/K$, the computational complexity of fitting our model is $\bigO(JN^3/K^2)$ for $J$ particles. SMC methods allows for parallel computation, as we can update the particles independently. Thus, we can reduce the computational complexity to fitting GP-MOE to be $\bigO(N^3/K^2)$. While the computational complexity of GP-MOE is considerably reduced from the typical $ \bigO(N^{3}) $ cost, the complexity still grows considerably as new data arrive. 

To mitigate this problem, we can adopt an approach where we subsample $ B \ll N_k $ observations uniformly without replacement within a mixture assignment to approximate the likelihood $ P(\y_{k} | \X_k, \btheta_{k}) $ by upweighting the likelihood by a power of $ N_{k}/B $. This stochastic approximation provides an estimate for the full likelihood and has been used successfully in other Bayesian inference algorithms \cite{Minsker:2014,SriCevDinDun2015}. 
\begin{equation}
\begin{aligned}
\bu_k^{(j)} = (u_1, \ldots , u_B) \sim \mbox{HyperGeometric}(B, \{i : z_i^{(j)}=k  \} ), \\
(\y_{\bu_k}, \X_{\bu_k}) = (y_u, \x_u : u \in \bu_k^{(j)}),  \\
P(\y_{\bu_k^{(j)}}| \X_{\bu_k^{(j)}}, \btheta_{k}^{(j)})\sim \mathcal{N} \left(0, \bSigma_{\theta_{k}^{(j)}} + \frac{N_k\sigma^2_{k}}{B}  \I\right).
%P(\y_{k} | \X_k, \btheta_{k}^{(j)}) \approx
\end{aligned}
\end{equation}
We use this stochastic approximation when the number of observations in a mixture, $ N_k^{(j)} $ exceeds $ B $, so that the complexity of fitting the GP-MOE does not increase when $  N_k^{(j)} > B$. With this stochastic approximation, the complexity of our model is $ \bigO(J \min\{N_k, B\}^{3} / K^{2}) $. In Table~\ref{table:complexity}, we compare the computational complexity of our GP-MOE method, POE (our method using only one particle), WISKI \cite{Stanton:Maddox:Delbridge:Wilson:2021} and OSVGP \cite{Bui:Nguyen:Turner:2017}. 
% have these methods been explained? have the acronyms been spelled out? Have the citations been mentioned? I don't recall any of them. essential to add all three.XXX done -mz

\begin{algorithm*}
	\caption{Online GP-MOE}\label{alg:bandit_online_update}
	\KwIn{New observation, $ (\x_i, y_i) $.}
		\tcc{Particle propagation.}
		\For{$ j = 1 , \ldots , J $ in parallel}{
			Sample $ z_{i}^{(j)}$ from $ P(z_{i}^{(j)} | \alpha^{(j)}, \X_{1:i-1})$, in Eq.~\ref{eqn:marg_t}\\			 
			Sample $ \btheta_{z_{i}}^{(j)} $ using the elliptical slice sampler from \cite{Murray:Adams:2010}.\\	
			Sample $ \alpha^{(j)} $	from the full conditional $ P(\alpha^{(j)} | \z_{1:i} )$ in Eq.~\ref{eqn:alpha_update}.\\
			Update particle weight $ \wtJ $ using Eq~\ref{eqn:weight_update}.
%			\begin{equation*}
%			%\wtJ = \wtminusJ \cdot \frac{ \prod_{k \in K^{+}} P(\y_k | \X_k, \z_{1:i}^{(j)}, \btheta_{k}^{(j)}) }{\prod_{k\in K^{\prime+}} P(\y_k^{\prime} | \X_k^{\prime}, \z_{1:i-1}^{(j)}, \btheta^{(j)}_{k}) } \cdot P( \x_{i} |z_{i}^{(j)}, \alpha^{(j)} )			
%			\wtJ := \wtminusJ \cdot \frac{  P\left( \y_{z_{i}^{(j)}} \left| \X_{z_{i}^{(j)}}, , \btheta_{z_{i}^{(j)}}^{(j)} \right. \right) }{ P\left(\y_{z_{i}^{(j)}}^{\prime} \left| \X_{z_{i}^{(j)}}^{\prime},  \btheta^{(j)}_{z_{i}^{(j)}}\right.\right) } \cdot P( \x_{i} |z_{i}^{(j)}, \alpha^{(j)} )			
%			\end{equation*}\\			
%			where
%			\begin{equation*}
%			\begin{aligned}
%			(\X_{k}^{\prime}, \y_{k}^{\prime}) = (\x_{\iprime}, y_{\iprime} , \iksum ),   (\X_{k},\y_{k}) = (\x_{i},y_{i},  i: z_{i}=k)
%			\end{aligned}
%			\end{equation*}			
		}
		Normalize weights, $\wtJ := \wtJ/\sum_{j=1}^{J}\wtJ$.\\
		\tcc{Particle resampling.}
		\If{$ N_{\mbox{eff}} < J/2 $}{
			Resample particles  $\left(\z^{(\jast)}_{1:i}, \btheta^{(\jast)}, \alpha^{(\jast)} \right)  $  from $\jast \sim \mbox{Multinomial}\left(J, \Ws \right)  $.\\
			Set $\wtJ := 1/J $ for $ j = 1 , \ldots , J $.
		}
	\KwOut{Particle weights $ \left(w_{i}^{(1)}, \ldots , w_{i}^{(J)}\right) $ and particles $ \left(\z^{(1:J)}_{1:t}, \btheta^{(1:J)}, \alpha^{(1:J)} \right) $.}
\end{algorithm*}

    \begin{algorithm}
	    \caption{GP-MOE Prediction}\label{alg:bandit_online_predict}
		\For{$ j = 1 , \ldots , J $ in parallel}{
			Predict new observations on particle $j$ with %(in practice, choose best $ k $ instead of average) 
			\begin{equation*}
			\begin{aligned}
			&p_k \propto N_k \cdot P(\x_{\ast} | z_{\ast}=k, \X_{k}, -)\\
			&P(y_{\ast}^{(j)} | \y, \X, \x_{\ast}, -)  = \sum_{k\in K^{+}} p_k \cdot P(y_{k\ast}^{(j)} | \y_{k}, \X_{k}, \x_{\ast},-)\\			
			\end{aligned}
			\end{equation*}
		} 
		Average predictions: $ P( \bar{y}_{\ast} | \y, \X, \x_{\ast}) = \sum_{j=1}^{J} \wtJ P(y_{\ast}^{(j)} | \y, \X, -) $

\end{algorithm}

\begin{table}
    \centering
    \begin{tabular}{ccc}
     GP & Sparse GP & WISKI \\  
    \midrule
     $\bigO(N^{3})$ & $\bigO(NM^2)$ & $ \bigO(M\log M) $ \vspace{1em}\\
     
     POE &\textbf{GP-MOE} & \textbf{Minibatched GP-MOE}\\
      \midrule
     $\bigO(N^3/K^2)$ & $\bigO(JN^3/K^2)$ & $\bigO(J\min\{N_k, B\}^{3}/K^2)$
    \end{tabular} 
      \caption{Comparison of inference complexity. $N$ is the number of data points, $K$ is the number of experts, $M$ is the number of inducing points, and $J$ is the number of particles.}\label{table:complexity}
\end{table}

\subsection{Optimization in a Bandit Setting with Online GP-MOE}
To motivate using the online GP-MOE for an applied problem, we show how this approach can be useful for a typical setting where GPs are popularly used--optimization. We can perform bandit Gaussian process optimization using our online GP-MOE method. The objective of this setting is that we want to optimize the function, $f$, using a mixture of GPs, where the arms of the bandit are indexed by $k$, corresponding to the kernel-specific hyperparameters of the mixture components. 

We update the model by first selecting a new test point $ \x_i $ using the UCB Thompson-Sampling from Eqn.~\ref{eqn:ts}. Then, we evaluate the point $ \x_i $ at $ f(\x_i) $ and update the model using Alg.~\ref{alg:bandit_online_update} at $(\x_i,f(\x_i))$. $ \x_i $ stochastically selects the arm, $ z_i $, which produces the best reward that corresponds to the marginal likelihood $P(\y_{k}|\X_{k},\btheta_k,\z_{1:i})$. Lastly, we can update the particle weight, $ \wtJ $, and the other model parameters, $ \btheta_k^{(j)} $ and $ \alpha^{(j)} $ after selecting the arm. We continue this procedure for $ N $ total function evaluations.

\begin{algorithm}
	\caption{GP-MOE Bandit Optimization with Thompson Sampling}\label{alg:bandit_opt}
	\For{$ i=1 , \ldots , N $}{		
		Sample points with $
		(y^{\ast}_1, \ldots , y^{\ast}_{N^{\ast}}) \sim P(\y^{\ast} | \X, \X^{\ast}, \y )$ using Alg.~\ref{alg:bandit_online_predict}.\\
		Select new point , $\x_i :=  \x_{\argmax_{i^{\ast}} (y^{\ast}_1, \ldots , y^{\ast}_{N^{\ast}})} .$ \\
%		where:
%		\begin{equation*}
%		\begin{aligned}
%		\m(\x_{\ast}) = \expt \left[ P( \bar{y}_{\ast} | \y, \X, \x_{\ast}) \right],\\
%		\bs(\x_{\ast}) =  \var \left(P( \bar{y}_{\ast} | \y, \X, \x_{\ast})\right). 
%		\end{aligned}
%		\end{equation*}
		Update model with $ (\x_i, f(\x_i)) $ using Alg.~\ref{alg:bandit_online_update}.
	}	
\end{algorithm}

	\section{Empirical Analyses for Online GP-MOE}\label{ref:experiments}
To demonstrate the ability of our algorithm to fit streaming non-stationary GPs, we apply our online GP-MOE to a collection of empirical time-series datasets that exhibit non-stationary behavior\footnote{The motorcycle dataset is available in the R package \texttt{VarReg}. The Brent, Canada CO$_2$, and Nile datasets are available here: \href{https://github.com/alan-turing-institute/TCPD}{\texttt{https://github.com/alan-turing-institute/TCPD}}. The EUR-USD dataset is available in the R package \texttt{priceR}.}. In our comparisons, we look at the following datasets: 1.) An accelerometer measurement of a motorcycle crash. 2.) The price of Brent crude oil. 3.) The annual water level of the Nile river data. 4.) The exchange rate between the Euro and the US Dollar. 5.) The annual carbon dioxide output in Canada. 6.) The heart rate of a hospital patient in the MIMIC-III data set \cite{Johnson:Pollard:2016}. We pre-process the data so that the inputs and outputs have zero-mean and unit variance.

% none of these acronyms make sense in light of the explanation.
We compare our method against three alternative approaches: a product-of-experts model (POE), which is a special case of our algorithm with only one particle, a sparse online GP method using the Woodbury identity and structured kernel interpolation (WISKI) \cite{Stanton:Maddox:Delbridge:Wilson:2021}, and an online sparse variational GP method\footnote{The implementation for OSVGP and WISKI is available here: \href{https://github.com/wjmaddox/online_gp}{\texttt{https://github.com/wjmaddox/online\_gp}}. Our code is available at \href{https://github.com/michaelzhang01/GPMOE}{\texttt{https://github.com/michaelzhang01/GPMOE}}.} (OSVGP) \cite{Bui:Nguyen:Turner:2017}. Our choice of kernel for each of these methods is the radial basis function (RBF):
\begin{equation}
\Sigma_{\theta}(\x,\x^{\prime}) = \exp\left\{ -\frac{\theta}{2}\sum_{d=1}^{D}(x_{d} - x_{d}^{\prime})^{2} \right\},
\end{equation}
for length scale parameter, $ \theta $.

In these experiments, we set the number of inducing points to be $50$ for each of the sparse methods. We evaluate our method when $ J $ is equal to $ 1 $ (equivalent to a POE), $ 100 $, and $ 500 $ particles. The OSVGP method requires a fixed number of optimization iterations to estimate the variational parameters, and the results are highly sensitive to this setting. To make the model settings comparable to the GP-MOE settings, we also set the optimizer iterations to $ 1 $, $ 100 $, and $ 500 $. In the GP-MOE model, we distribute the inference of each particle over $16$ cores on a shared memory process using OpenMP. In this setting, we run an online prediction experiment where we initialize the model using the first observation in the time-series data set. Then, we sequentially predict the subsequent observation and update the model with the next data point.

Our method generally performs better than the competing online GP methods (Tables~\ref{table:results_LL} and~\ref{table:results_mse}). We broadly observe that the GP-MOE performs better than POE because we can integrate over the space of partitions and, thus, we will better capture the predictive uncertainty. However, WISKI is undoubtedly the fastest method as the computational complexity is constant with respect to the number of observations. But because these data sets exhibit non-stationarity, WISKI is not able to handle changes in the kernel behavior (like heteroscedasticity, for example) and therefore performs poorly in terms of MSE and log likelihood in these experiments. The GP-MOE methods perform comparatively to the OSVGP, which only uses a single GP, in terms of wall time (Table~\ref{table:results_wall_time}).

The performance of the OSVGP strongly depends on the number of optimizer iterations, and OSVGP has poor performance when only one iteration is used as opposed to 500 iterations. OSVGP often runs into numerical stability issues relating to calculating Cholesky decompositions in the sparse variational GP \cite{Stanton:Maddox:Delbridge:Wilson:2021}. Like WISKI, the OSVGP can sometimes obtain adequate point estimates for the posterior mean but generally mischaracterizes the posterior predictive variance in these non-stationary examples. While OSVGP can somewhat capture the non-stationary behavior as the hyperparameters update as new data arrive, the posterior predictive variance is generally wide compared to GP-MOE as evidenced by the superior performance of GP-MOE with respect to the predictive log likelihood.

For the motorcycle, Brent, Canadian carbon dioxide, and heart rate datasets, the GP-MOE performs the best in terms of predictive mean squared error and log likelihood. In the Nile river data set, the OSVGP performs the best in terms of online predictive log likelihood. This could be because the only non-stationary component of the Nile river data set is the mean value, which we assume to be constant at all values of $ \x $ in GP-MOE. The OSVGP and WISKI obtain the best MSE and log likelihood results on the EUR-USD data sets as well, which exhibits only time varying noise, zero mean, and stationary length-scale for the entire duration of the time-series data. Here, OSVGP and WISKI produce wider noise estimates than GP-MOE. However, a stochastic volatility model would perhaps be a better fit for this type of data than a mixture of GPs. The other datasets exhibit time-varying noise terms and length scales, which our mixture of GPs can capture, and thus the GP-MOE exhibits superior performance.

\begin{figure*}
	\centering
	\subfloat[GP-MOE]{\includegraphics[width=.25\linewidth]{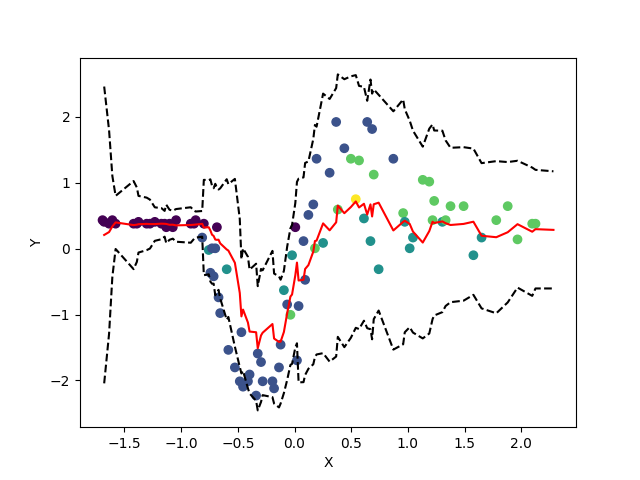}}
	\subfloat[POE]{\includegraphics[width=.25\linewidth]{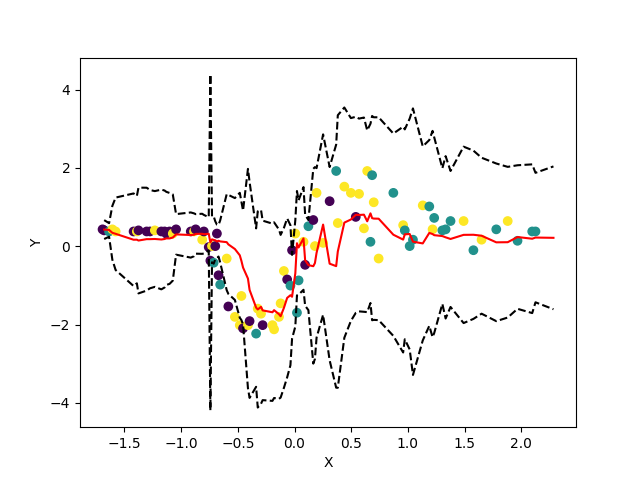}}
	\subfloat[WISKI]{\includegraphics[width=.25\linewidth]{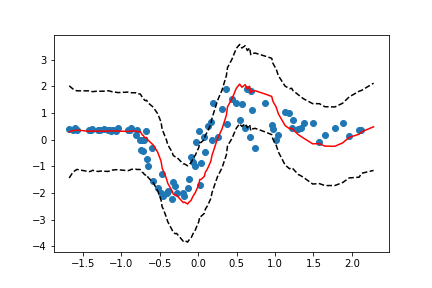}}
	\subfloat[OSVGP]{\includegraphics[width=.25\linewidth]{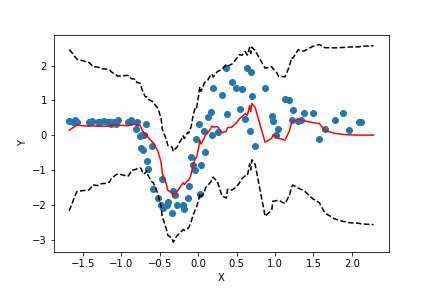}}
	\caption{Online posterior predictive mean (plotted with solid red lines) and 95\% credible intervals (plotted with dashed black lines) for the motorcycle dataset. The color of the data points in these figures for GP-MOE and POE represent the mixture assignment of that observation for the particle with the highest weight. $ N=94 $.}
	\label{fig:motorcycle_compare}
\end{figure*}

\begin{figure*}
	\centering
	\subfloat[GP-MOE]{\includegraphics[width=.25\linewidth]{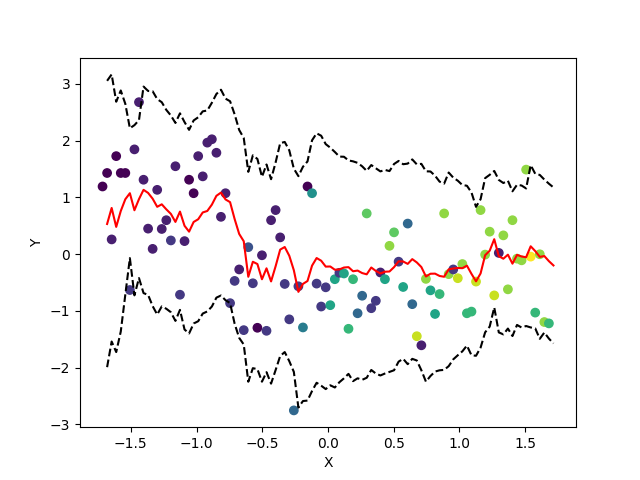}}
	\subfloat[POE]{\includegraphics[width=.25\linewidth]{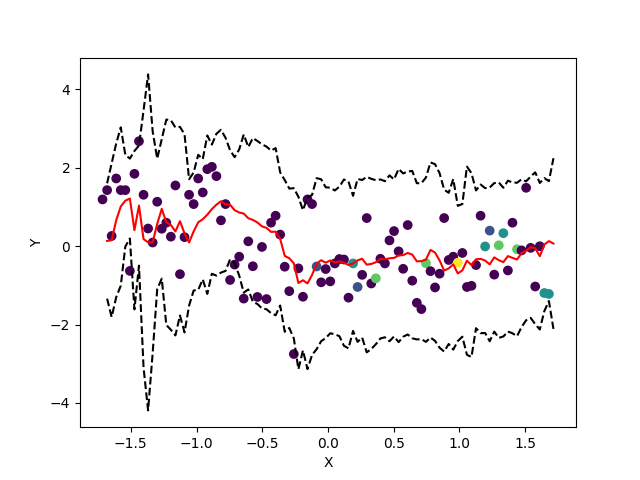}}
	\subfloat[WISKI]{\includegraphics[width=.25\linewidth]{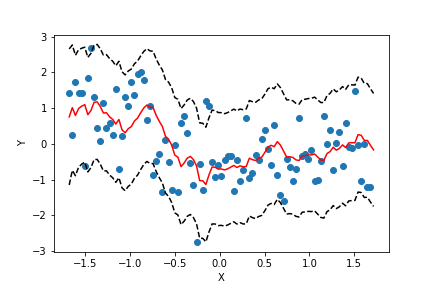}}
	\subfloat[OSVGP]{\includegraphics[width=.25\linewidth]{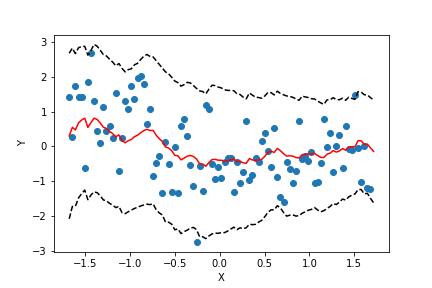}}
	\caption{Online posterior predictive mean (plotted with solid red lines) and 95\% credible intervals (plotted with dashed black lines) for the Nile river dataset. The color of the data points in these figures for GP-MOE and POE represent the mixture assignment of that observation for the particle with the highest weight. $ N=100 $.}
	\label{fig:nile_compare}
\end{figure*}

\begin{figure*}
	\centering
	\subfloat[GP-MOE]{\includegraphics[width=.25\linewidth]{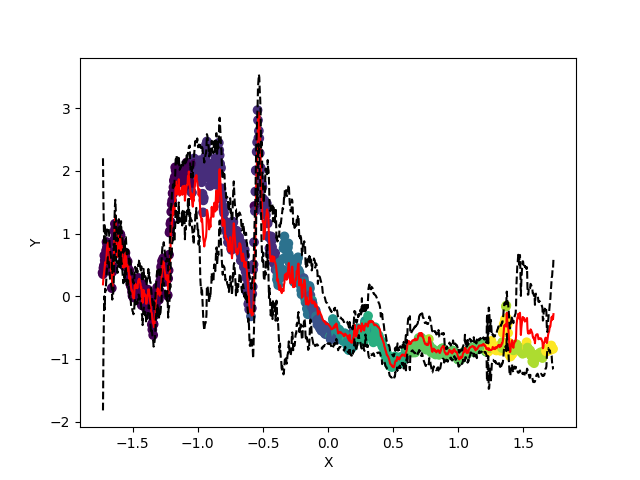}}
	\subfloat[POE]{\includegraphics[width=.25\linewidth]{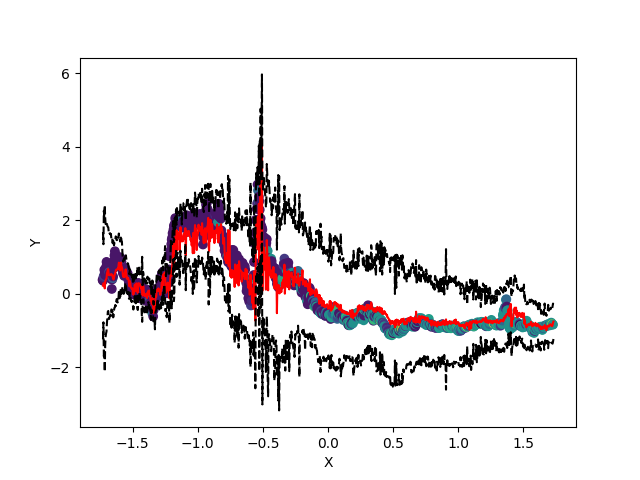}}
	\subfloat[WISKI]{\includegraphics[width=.25\linewidth]{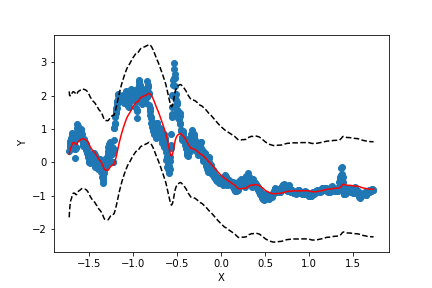}}
	\subfloat[OSVGP]{\includegraphics[width=.25\linewidth]{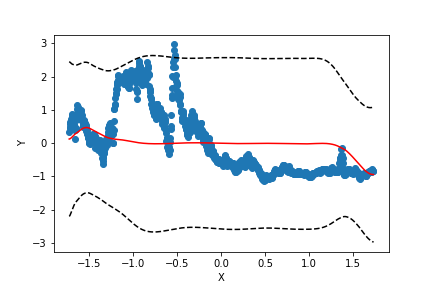}}
	\caption{Online posterior predictive mean (plotted with solid red lines) and 95\% credible intervals (plotted with dashed black lines) for the Brent crude oil dataset. The color of the data points in these figures for GP-MOE and POE represent the mixture assignment of that observation for the particle with the highest weight. $ N=1025 $.}
	\label{fig:brent_compare}
\end{figure*}

\begin{figure*}
	\centering
	\subfloat[GP-MOE]{\includegraphics[width=.25\linewidth]{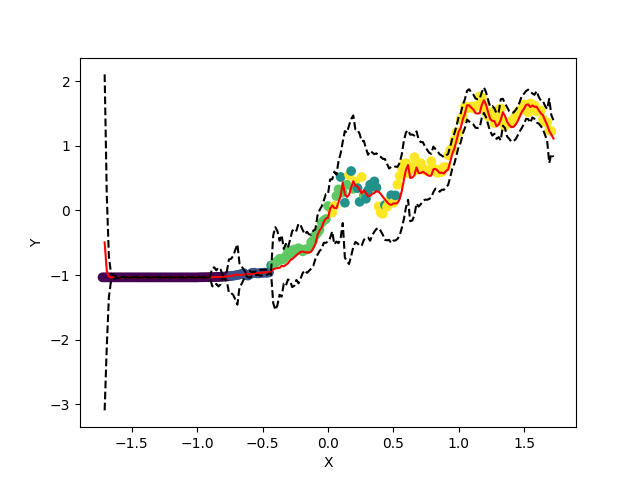}}
	\subfloat[POE]{\includegraphics[width=.25\linewidth]{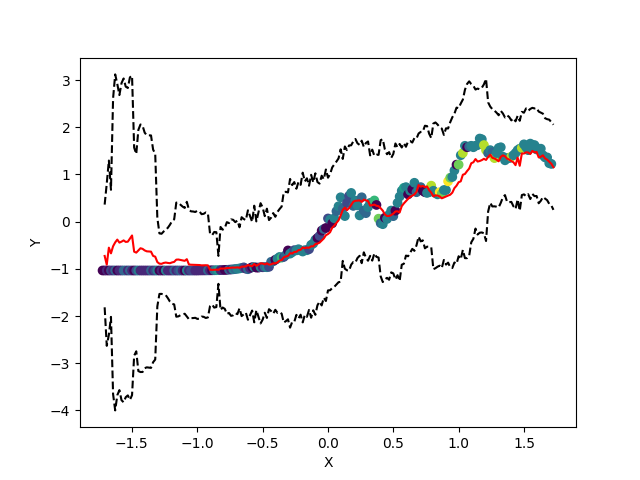}}
	\subfloat[WISKI]{\includegraphics[width=.25\linewidth]{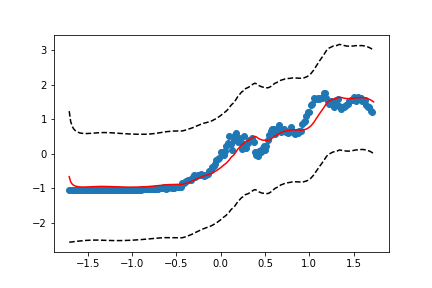}}	\subfloat[OSVGP]{\includegraphics[width=.25\linewidth]{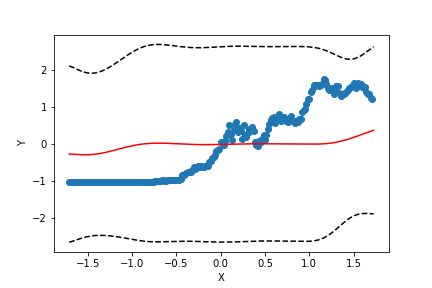}}
	\caption{Online posterior predictive mean (plotted with solid red lines) and 95\% credible intervals (plotted with dashed black lines) for the Canadian CO$_2$ dataset. The color of the data points in these figures for GP-MOE and POE represent the mixture assignment of that observation for the particle with the highest weight. $ N=215 $.}
	\label{fig:canada}
\end{figure*}

\begin{figure*}
	\centering
	\subfloat[GP-MOE]{\includegraphics[width=.25\linewidth]{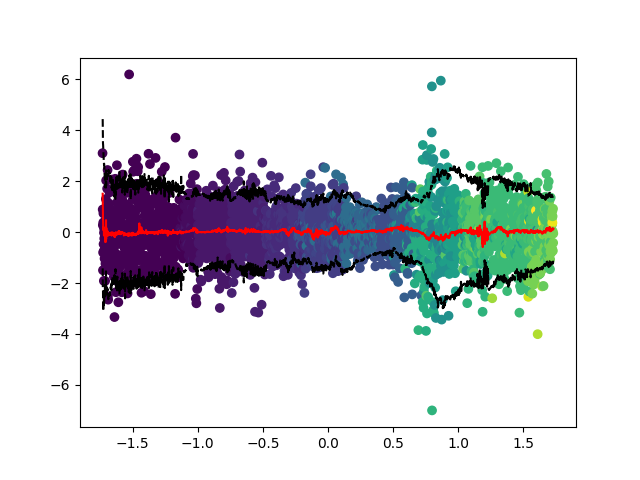}}
	\subfloat[POE]{\includegraphics[width=.25\linewidth]{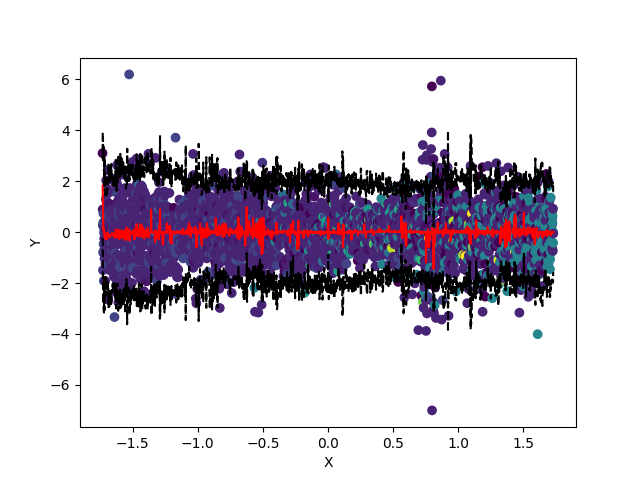}}
	\subfloat[WISKI]{\includegraphics[width=.25\linewidth]{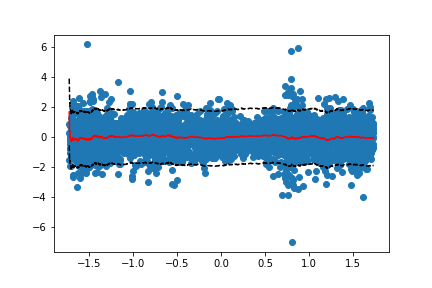}}
	\subfloat[OSVGP]{\includegraphics[width=.25\linewidth]{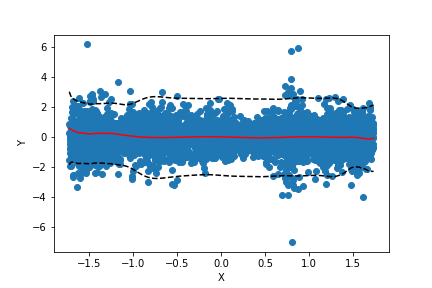}}
	\caption{Online posterior predictive mean (plotted with solid red lines) and 95\% credible intervals (plotted with dashed black lines) for the USD-EUR exchange rate dataset. The color of the data points in these figures for GP-MOE and POE represent the mixture assignment of that observation for the particle with the highest weight. $ N=3139 $.}
	\label{fig:usd_compare}
\end{figure*}

The online predictive performance of each of these online GP models show that WISKI and OSVGP produce overly wide credible intervals in comparison to GP-MOE and POE (Figures~\ref{fig:motorcycle_compare}-\ref{fig:usd_compare}). In some instances, WISKI and OSVGP produce inaccurate predictive means in the case of the motorcycle, Brent, and Canada data sets. In the GP-MOE and POE results, we have colored the observations based on the mixture assignments of the largest particle. We can see from these plots that the mixture assignments largely correspond to changes in the underlying functional behavior. For example in the plot for the motorcycle data set, we observe an initial low-noise regime and a subsequent high-noise regime. The GP-MOE assigns data to different mixtures in these separate parts of the time series where the noise and length scale behavior changes (Figures~\ref{fig:motorcycle_compare}-\ref{fig:usd_compare}). In contrast, stationary methods like WISKI or OSVGP model the motorcycle data with a constant noise term across time, and we can see that the initial low-noise regime of the data is modeled with an extremely large predictive 95\% credible interval. 

% For the GP-MOE, we can see that the observations are assigned to different mixtures when the underlying function changes behavior. <- this should be in the caption and not here. I also dont entirely undrstand the point you are trying to make here.

%, \ref{fig:nile_compare}, \ref{fig:brent_compare}, \ref{fig:usd_compare}

\begin{figure*}
	\centering
	\subfloat[GP-MOE]{\includegraphics[width=.25\linewidth]{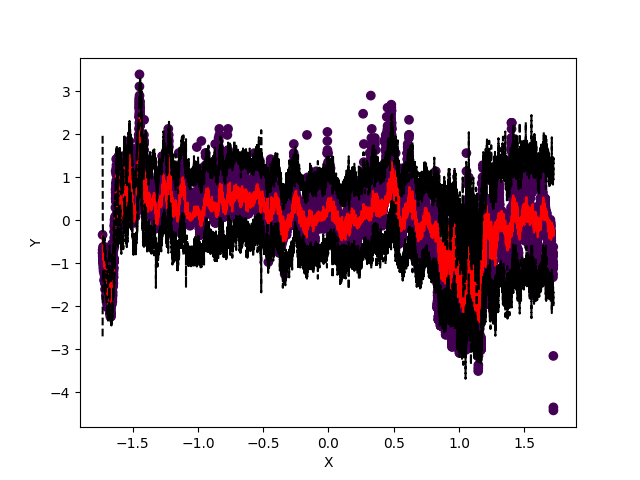}}
	\subfloat[POE]{\includegraphics[width=.25\linewidth]{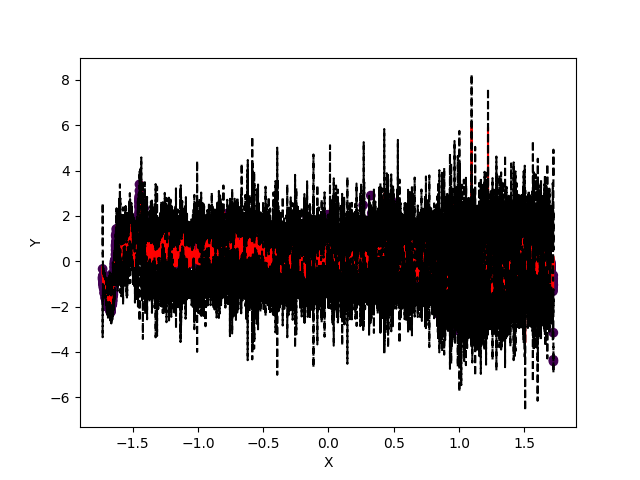}}
	\subfloat[WISKI]{\includegraphics[width=.25\linewidth]{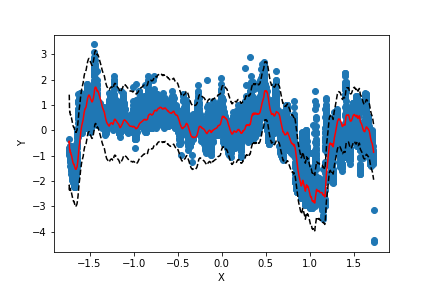}}
	\subfloat[OSVGP]{\includegraphics[width=.25\linewidth]{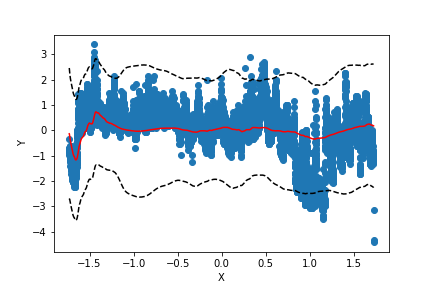}}
	\caption{Online posterior predictive mean (plotted with solid red lines) and 95\% credible intervals (plotted with dashed black lines) for the heart rate dataset. The color of the data points in these figures for GP-MOE and POE represent the mixture assignment of that observation for the particle with the highest weight. $ N=10000 $.}
	\label{fig:hr}
\end{figure*}

\subsection{Hyperparameter Settings}
Our approach has important hyperparameters that can affect the performance of our online GP-MOE algorithm. Specifically, these important hyperparameters are the number of particles and the minibatch size. The authors of the IS-MOE method provides an in-depth examination of how changing these hyperparameters affect the predictive performance in terms of test set log likelihood and mean squared error \cite{Zhang:Williamson:2017}. Increasing the number of importance samples, $J$, and minibatch size $ B $ of the algorithm will improve model performance and increase computation time, while increasing the number of experts, $K$, will decrease both computation time and model performance (as $K$ increases, the covariance matrix becomes diagonally dominant, and is unable to capture input correlation). However, there are intermediate settings for each parameter that can drastically reduce computation time without a decrease in predictive model performance. 

\begin{figure*}
	\centering
	\includegraphics[width=0.5\linewidth]{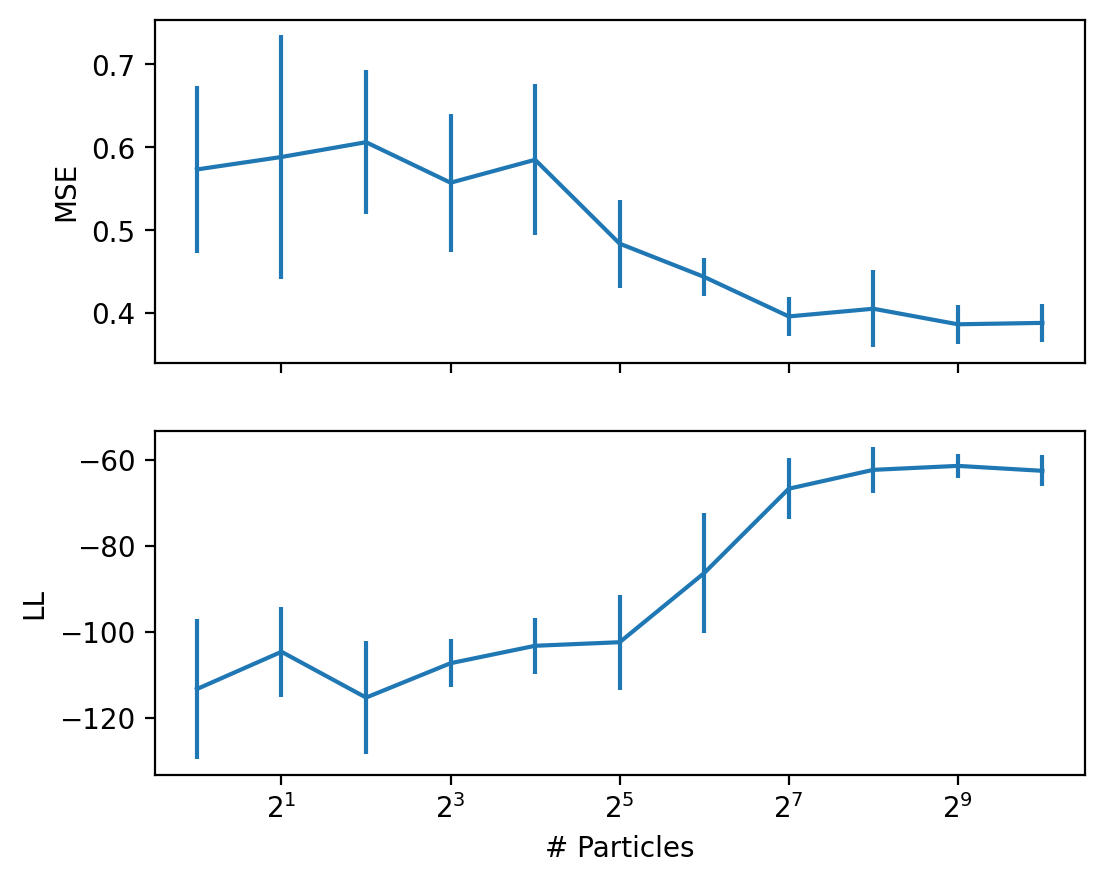}\includegraphics[width=0.5\linewidth]{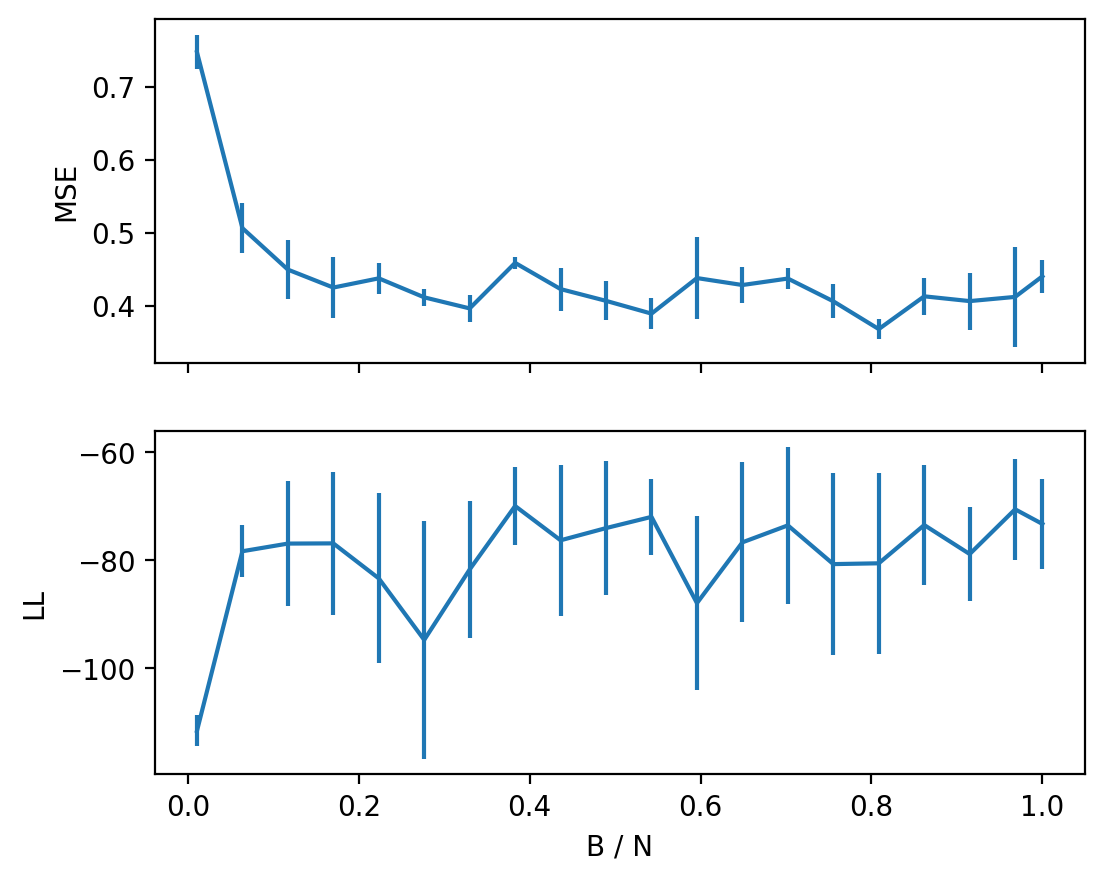}
	\caption{Left: Performance of GP-MOE with increasing values of $ J $ on the motorcycle dataset. Right: Performance of GP-MOE with increasing values of $ B $ on the motorcycle dataset.}
	\label{fig:j_motorcycle}
\end{figure*}

We can look at the effect of increasing $ J $ and $ B $ on the online predictive MSE and log likelihood (Figure~\ref{fig:j_motorcycle}). 
%(Figures~\ref{fig:j_motorcycle} and~\ref{fig:mb_motorcycle}). 
Unsurprisingly, increasing $ J $ and $ B $ will improve the predictive performance of GP-MOE. However, there is a point at which the performance of GP-MOE will plateau with respect to an increasing $ J $ and $ B $, indicating that there are lower settings for which we can obtain accurate performance.

%\begin{figure*}
%	\centering
%	\includegraphics[width=1.0\linewidth]{figs/mb_motorcycle}
%	\caption{}
%	\label{fig:mb_motorcycle}
%\end{figure*}
%
%
%\begin{figure}
%	\centering
%	\includegraphics[width=1.0\linewidth]{figs/J_motorcycle}%\includegraphics[width=0.4\linewidth]{figs/mb_motorcycle}
%	\caption{Performance of GP-MOE with increasing values of $ J $ on the motorcycle dataset. }
%	\label{fig:j_motorcycle}
%\end{figure}
%
%\begin{figure}
%	\centering
%	\includegraphics[width=1.0\linewidth]{figs/mb_motorcycle}
%	\caption{Performance of GP-MOE with increasing values of $ B $ on the motorcycle dataset.}
%	\label{fig:mb_motorcycle}
%\end{figure}

\subsection{Bandit Optimization}
\begin{figure*}
	\centering
	\subfloat[Branin-Hoo]{\includegraphics[width=.25\linewidth]{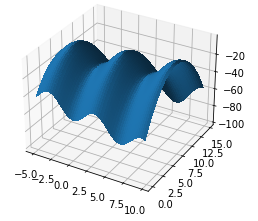}}
	\subfloat[Parsopoulos]{\includegraphics[width=.25\linewidth]{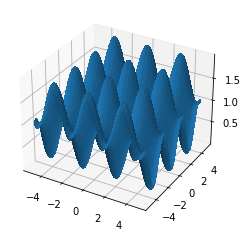}}		
	\subfloat[Styblinski-Tang]{\includegraphics[width=.25\linewidth]{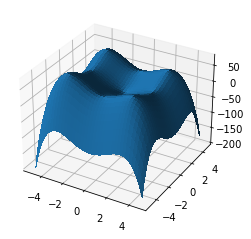}}	
	\subfloat[Ursem]{\includegraphics[width=.25\linewidth]{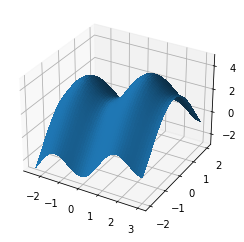}}	
	\caption{Test functions for the bandit optimization experiments.}\label{fig:function_opt}
\end{figure*}

Next, we use our GP-MOE in a bandit optimization setting using Thompson sampling. Here, we are maximizing four functions typically used to evaluate optimization algorithms: the Branin-Hoo function,  the Syblinski-Tang function, the Parsopoulos function, and the Ursem function (plotted in Figure~\ref{fig:function_opt}). We choose these functions as they are multimodal, and we want to examine how well the competing GP optimization methods both explore and exploit the target function. 

We compared GP-MOE, POE, WISKI, OSVGP and GP-TS on these four optimization tasks, setting $ J=100 $ for the GP-MOE, and the number of inducing points to be $ 50 $ for the sparse methods. We run each method for $ N=500 $ iterations. In bandit optimization, we focus on two results: Minimizing the mean average regret (MAR) and obtaining the maximum value of the function $ \max_{\x} f(\x) $. GP-MOE and POE typically obtain the highest (best) maximum value, while the stationary methods (especially the basic GP-TS) obtain better MAR (Table~\ref{table:results_opt}). 

This suggests that basic online GP approaches are liable to only explore a suboptimal mode. While the bandit can reliably estimate the neighborhood surrounding this mode, thus resulting in a low MAR, it cannot actually find a better optimum compared to GP-MOE and POE. The only setting where a stationary method can compete with GP-MOE and POE is for the Parsopoulos function where GP-TS performs comparably to GP-MOE and POE. Though the POE can attain a maximum function value on par with GP-MOE, we see that the GP-MOE produces a maximum value with typically lower variance than the POE, which is an effect we would expect to see in an ensemble method like SMC. In these optimization experiments, we see that POE has the lowest CPU wall time. Thus, for an optimization task, like the ones analysed in this section, the optimal setting may be one where the number of particles, $ J $, is rather low in order to take advantage of the variance minimization of SMC while maintaining the speed of the POE.

\begin{table*}
	\centering
	\resizebox{\textwidth}{!}{
		\begin{tabular}{l|llllll}
			& Motorcycle & Nile & Brent & Canada & EUR-USD & Heart Rate \\
			\hline
			GP-MOE 500 & \textbf{-63.686 (2.370)} & -144.397 (2.447) & \textbf{266.563 (45.322)} & \textbf{305.852 (14.763)} & -4585.758 (20.097) & \textbf{-8328.856 (31.848)} \\
			GP-MOE 100 & -72.157 (3.841) & -147.531 (4.369) & 120.428 (73.548) & 261.257 (37.422)   & -4539.573 (42.731) & -8790.210 (32.288) \\
			POE & -114.665 (11.187) & -142.832 (1.513) & -456.628 (17.927) & -114.958 (12.374)& -4538.173 (25.029) & -12023.802 (158.452) \\
			WISKI & -112.467 (0.000) & -127.998 (0.000) & -800.852 (0.000) & -152.242 (0.000) & \textbf{-4482.185 (0.000)} & -10261.351 (0.000) \\
			OSVGP 500 & -99.523 (3.019) & \textbf{-127.289 (0.157)} & -1250.734 (142.780) & 43.021 (5.292)  & -4766.513 (43.271) & -14307.737 (278.164) \\
			OSVGP 100 & -125.862 (0.306) & -138.537 (0.057) & -731.435 (156.499) & -230.525 (0.893) & -4494.476 (0.244) & -14550.117 (107.419) \\
			OSVGP 1 & -135.776 (0.100) & -145.048 (0.116) & -1438.428 (3.493) & -313.022 (0.296) & -4676.530 (5.742) & -13586.578 (145.657) 
		\end{tabular}
	}
	\caption{Online predictive log likelihood over five trials. One standard error reported in parentheses.}\label{table:results_LL}
\end{table*}

\begin{table*}
	\centering
	\resizebox{\textwidth}{!}{
		\begin{tabular}{l|llllll}
			& Motorcycle & Nile & Brent & Canada & EUR-USD & Heart Rate \\
			\hline
			GP-MOE 500 & \textbf{0.389 (0.007)} & \textbf{0.722 (0.003)} & \textbf{0.049 (0.006)} & 0.019 (0.002) & 1.010 (0.002) & \textbf{0.379 (0.002)} \\
			GP-MOE 100 & 0.417 (0.019) & 0.740 (0.005) & 0.061 (0.008) & \textbf{0.018 (0.002)}  & 1.006 (0.002) & 0.438 (0.002) \\
			POE & 0.479 (0.038) &  0.807 (0.017) & 0.123 (0.007) & 0.102 (0.019) & 1.016 (0.002) & 0.677 (0.006) \\
			WISKI & 0.631 (0.000) & 0.767 (0.000) & 0.177 (0.000) & 0.048 (0.000) & 1.007 (0.000) & 0.408 (0.000) \\
			OSVGP 500 & 0.413 (0.028) & 0.765 (0.003) & 0.922 (0.047) & 0.028 (0.001) & 1.036 (0.013) & 0.939 (0.011)  \\
			OSVGP 100 & 0.802 (0.004) & 0.852 (0.001) & 0.444 (0.050) & 0.366 (0.004) & \textbf{1.003 (0.000)} & 0.935 (0.005) \\
			OSVGP 1 & 0.986 (0.001) & 0.934 (0.001) & 0.916 (0.003) & 0.929 (0.005) & 1.017 (0.002) & 0.812 (0.016)
		\end{tabular}
	}
	\caption{Online predictive mean squared error over five trials. One standard error reported in parentheses.}\label{table:results_mse}
\end{table*}

\begin{table*}
	\centering
	\resizebox{\textwidth}{!}{
	\begin{tabular}{l|llllll}
		& Motorcycle & Nile & Brent & Canada & EUR-USD & Heart Rate \\
		\hline
		GP-MOE 500 & 608.442 (5.507) & 235.979 (6.036) & 12503.964 (1133.177) & 2956.015 (112.812) & 67034.274 (15869.414) & 34015.384 (286.470) \\
		GP-MOE 100 & 203.195 (2.664) & 69.393 (0.808) & 4505.307 (508.852) & 1077.938 (39.106) & 11200.608 (2385.012) & 8896.295 (20.048) \\
		POE & 85.524 (1.488) & 28.604 (0.529) & 1053.192 (98.926) & 324.706 (19.900) & 1597.147 (398.397) & 1094.152 (9.027) \\
		WISKI & \textbf{1.772 (0.363)} & \textbf{1.347 (0.095)} & \textbf{16.297 (3.233)} & \textbf{3.429 (0.170)} & \textbf{50.901 (6.921)} & 463.025 (33.449) \\
		OSVGP 500 & 1693.370 (182.273) & 1648.170 (99.036) & 19176.419 (2168.529) & 3908.914 (239.521) & 57412.117 (1644.124) & 193332.795 (14744.607) \\
		OSVGP 100 & 323.504 (74.718) & 347.613 (40.670) & 3819.830 (350.382) & 830.059 (118.029) & 10829.119 (1571.008) & 41138.089 (2890.219) \\
		OSVGP 1 & 2.729 (0.631) & 3.156 (0.543) & 33.614 (3.736) & 5.929 (0.254) & 83.713 (10.725) & \textbf{442.911 (22.729)}
	\end{tabular}
	}
	\caption{CPU wall time in seconds for online prediction over five trials. One standard error reported in parentheses.}\label{table:results_wall_time}
\end{table*}

\begin{table*}
	\centering
	\resizebox{\textwidth}{!}{
	\begin{tabular}{l|llllllll}
		& \multicolumn{2}{l}{Branin-Hoo} & \multicolumn{2}{l}{Parsopoulos} & \multicolumn{2}{l}{Styblinski-Tang} & Ursem &  \\
		& MAR & $\max f(\x)$ & MAR & $\max f(\x)$ & MAR & $\max f(\x)$ & MAR & $\max f(\x)$ \\ \hline
		MOE & 10.64 (5.54) & -0.41 (0.01) & \textbf{0.107 (0.005)} & \textbf{1.999 (0.001)} & 5.65 (1.38) & \textbf{78.06 (0.19)} & 0.63 (0.15) & 6.336 (0.029) \\
		POE & 23.85 (2.14) & \textbf{-0.42 (0.02)} & 0.202 (0.030) & \textbf{1.999 (0.001)} & 12.84 (1.87) & 78.05 (0.21) & 0.98 (0.19) & \textbf{6.342 (0.034)} \\
		WISKI & 2.50 (0.12) & -19.86 (0.04) & 0.418 (0.011) & 1.996 (0.003) & 0.418 (0.011) & 1.996 (0.003) & 0.42 (0.02) & 5.338 (0.082) \\
		OSVGP & 56.68 (0.59) & -20.35 (0.30) & 0.952 (0.030) & 1.985 (0.009) & 0.952 (0.030) & 1.985 (0.009) & 2.06 (0.07) & 5.266 (0.056) \\
		GP-TS & \textbf{0.96 (0.15)} & -19.93 (0.08) & 0.351 (0.038) & \textbf{1.999 (0.001)} & \textbf{0.351 (0.038)} & 1.999 (0.001) & \textbf{0.34 (0.01)} & 5.374 (0.020)
	\end{tabular}
	}
	\caption{Mean absolute regret and maximum function value for optimization experiments over five trials for 500 iterations. One standard error reported in parentheses.}\label{table:results_opt}
\end{table*}

\begin{table*}
	\centering
\begin{tabular}{l|llll}
	& Branin-Hoo            & Parsopoulos           & Styblinski-Tang       & Ursem                 \\ \hline
	MOE   & 1100.41 (624.45)      & 775.88 (180.24)       & 1154.05 (738.10)      & 1278.11 (299.29)      \\
	POE   & \textbf{21.53 (1.58)} & \textbf{25.78 (0.55)} & \textbf{21.88 (0.86)} & \textbf{22.25 (0.81)} \\
	WISKI & 51.90 (5.98)          & 44.78 (7.59)          & 56.00 (31.70)         & 48.95 (8.81)          \\
	OSVGP & 1346.01 (60.04)       & 1216.63 (167.58)      & 1546.88 (974.74)      & 1227.35 (116.67)      \\
	GP-TS & 1513.50 (187.90)      & 1161.05 (517.87)      & 949.81 (324.08)       & 779.65 (50.75)       
\end{tabular}
	\caption{CPU wall time for optimization experiments over five trials for 500 iterations. One standard error reported in parentheses.}\label{table:cpu_opt}
\end{table*}

%\begin{figure}
%	\centering
%	\includegraphics[width=0.95\linewidth]{figs/moe100_uk_pollution_seed0.png}
%	\caption{}
%	\label{fig:ukair}
%\end{figure}
	\section{Conclusion and Future Directions}\label{sec:conclusion}
In this paper, we introduced an online inference algorithm for fitting mixtures of Gaussian processes that can perform online estimation of non-stationary functions. We show that we can apply this mixture of GP experts in an optimization setting. In these bandit optimization experiments, we observe that GP-MOE finds the highest maximum value compared to the competing methods, and can generally produce better performance in terms of minimizing regret. 

For future work, we are interested in extending this method in a few different domains: First, we want to implement additional features that may improve the performance of our online method with regards to the sequential Monte Carlo sampler. Among these, we want to investigate incorporating retrospective sampling of the mixture assignments, $ \z $, as we only sample these indicators once when the $ i $-th observation arrives. Retrospective sampling could lead to better predictive performance in subsequent observations, though the additional computational cost could be prohibitive for extremely large data sets. 

Moreover, we are interested in introducing control variates in conjunction with the SMC sampler in order to reduce the variance of the Monte Carlo estimates. Basic Monte Carlo methods tend to exhibit high estimator variance. But, if we have some random variables correlated with our estimator whose expectation is known, then we can reduce the variance by a considerable amount. We seek to investigate these possible extensions in future work.

Next, we are interested in applying our online mixture of expert approach for modeling patient vital signs in a hospital setting. Gaussian processes have enjoyed notable success in the health-care domain, for example, in predicting sepsis in hospital patients \cite{Futoma:Hariharan:Heller:2017,Futoma:2017}, and in jointly modeling patients' vital signals through multi-output GPs \cite{Cheng:2020}. In future research, we will to extend our approach to multi-output GP models, and implement kernel functions customized for health care scenarios. By combining fast inference with flexible modeling, these approaches will have a profound impact in real-time monitoring and decision-making in patient health.

	\bibliographystyle{IEEEtran}
	\bibliography{online_gp}
\end{document}